# Elucidating Discrepancy in Explanations of Predictive Models Developed using EMR

Aida BRANKOVIC [a,1], Wenjie HUANG [c,2], David COOK [b,a,3], Sankalp KHANNA [a,4], Konstanty BIALKOWSKI [c,5]

[a] *CSIRO Australian e-Health Research Centre, Brisbane, QLD 4029, Australia*
[b] *Intensive Care Unit, Princess Alexandra Hospital, Brisbane, QLD 4102, Australia*
[c] *The University of Queensland, Brisbane, QLD, Australia*

ORCiD ID: Aida Brankovic https://orcid.org/0000-0001-7978-575X., David Cook https://orcid.org/0000-0001-9761- 0179, Sankalp Khanna https://orcid.org/0000-0001-6125-8871, Konstanty Bialkowski https://orcid.org/ 0000-0002-3461-0371

**Abstract.** The lack of transparency and explainability hinders the clinical adoption of Machine learning (ML) algorithms. While explainable artificial intelligence (XAI) methods have been proposed, little research has focused on the agreement between these methods and expert clinical knowledge. This study applies current state-of-the-art explainability methods to clinical decision support algorithms developed for Electronic Medical Records (EMR) data to analyse the concordance between these factors and discusses causes for identified discrepancies from a clinical and technical perspective. Important factors for achieving trustworthy XAI solutions for clinical decision support are also discussed.

## 1. Introduction

Machine Learning (ML) based tools have the potential to significantly impact health and healthcare delivery, yet these methods are often "black-box" in nature. In this context, it means the influential factors and how changes to these observable inputs can modify the results of the analysis are obscure or unknown. This is an important deficiency because understanding and being able to modify and reliably ("predictably") improve outcomes of a complex process that is modelled is the key to causing desired effects.

In addition to clinical ethics which imposes fundamental working principles on the tools and devices used for treatments, this creates a barrier to their adoption for clinical decision-making, as the lack of transparency, reduces their trustworthiness [1]. At the very least, an understanding of the methods should allow users to detect when they are not working or being used outside the purpose for which they were designed.

To understand decision of "black-box" ML models, a number of explainable artificial intelligence (XAI) methods have been proposed in the literature [1]. These can be categorised into gradient-based methods (e.g. SmoothGrad [2], Integrated Gradients [3], Deep Taylor Decomposition (DTD) [4]), Layer-wise propagation [5], and perturbation-based methods (e.g. LIME [6], Shap [7]). However, there is little understanding of how correct or relevant each of these is, or what level of agreement exists between results obtained with different methods.

---

[1] Corresponding Author: Aida Brankovic, aida.brankovic@csiro.au.

From clinicians' view, knowing the subset of features driving the model outcome is crucial as it allows them to compare the model decision to their clinical judgment, especially in case of a discrepancy [8]. A recent study [9] reports disagreement between the explanations generated by various popular explanation methods. This variability implies that some generated explanations (global and individual) could be misleading and possibly cause wrong decisions with dire consequences. Hence, it is critical to understand the accordance level between the explanations produced by state-of-the-art methods and domain expert knowledge.

To tackle these problems, this paper presents the results of a quantitative analysis of global explanations performed on two EMR datasets. The aim is to elucidate the agreement between the methods and with domain expert knowledge and to identify and discuss disaccord causes as a prerequisite for establishing the trustworthiness criteria for risk prediction models and their explanations in healthcare settings.

## 2. Methods

*2.1. Study cases*

In this study, two heath applications using two different datasets sourced from two metropolitan hospitals in Queensland, Australia (see section 2.2) are used. The first of these involves predicting patient readmission in a paediatric hospital and the second involves anticipating patient deterioration events in an acute adult inpatient ward setting. Ethics were obtained from Metro South Human Research Ethics Committee (Ref HREC/18/QPAH/525 and Ref HREC/16/QPAH/217).

*2.2. Data, predictors, and response variables*

The readmission study sourced data from Queensland Children's Hospital in Brisbane, Australia. De-identified inpatient paediatric administrative data from 1/1/2015 to 31/12/2018 were used for modelling and explainer development. Predictors included demographics, patient stay history as well as in-hospital medications pathology data. Readmission within the next 30 days (RA30) was the predicted outcome. The patient deterioration study sourced data from Princess Alexandra Hospital in Brisbane, Australia. De-identified adult inpatient data from 1/1/2016 to 31/1/2018 were used for modelling and explainer development. Predictors included Vital signs (Systolic and Diastolic Blood Pressure, Mean Arterial Pressure, Heart Rate, Temperature, Respiratory Rate, Oxygen Saturation (SpO2), Oxygen flow rate, level of consciousness (AVPU), and demographic information (age, gender, length of stay (LOS)). A red flag deterioration alert in the next 8 hours was the predicted outcome. For more information related to both datasets, interested readers are referred to published studies [10,11].

*2.3. Predictive models*

To explore the agreement between the features obtained by different methods three modelling approaches were selected, each as a representative of one modelling paradigm: regression, conventional ML, and Deep Neural Network (DNN). Logistic regression (LR) with l1 regularization (L1), often referred to as LASSO, is chosen because it is the most accepted predictive model. Unlike ML models which are generally more complex, it allows insight into the model structure and as such is the most trustworthy and accepted model among clinicians. XGB was considered as representative of conventional ML with a proven record of high performance when dealing with large and complex health data. Model parameters for L1 and XGB were employed as reported in [10,11]. A DNN model with convolution layers (CNN) that allows the extraction of new features was deployed as a representative of NN-based models.

*2.4. Explanations*

Shap and DTD explainers were selected owing to their popularity for explaining predictive models in healthcare and constructed on top of the ML models developed. Due to space limitations, this work does not present a comparison of any individual patients. Instead, to gain generalized insight into feature agreement and relevant features that underpinned decisions of predictive models, we considered the sum of explanations obtained for each patient individually across the features on the whole datasets (i.e. global explanations) and used them for this analysis.

*2.5. Measuring Agreement*

To measure agreement, we compared top features and their rankings obtained by different explanation methods using two metrics defined by Krishna et al. [11]: *i) Feature agreement (FA)* represents the fraction of common features between the sets of top-n features obtained by two explanation methods. *ii) Rank Agreement (RA)* represents the fraction of features with the same ranking in the sets of top-n features obtained by two explanation methods. We also focus on comparing the top features suggested by each approach with the expert-suggested predictors in each of the studies. The importance of focusing on these has also been identified as being critical by clinicians [8].

**3. Results**

Feature and rank agreement metrics computed for two considered datasets and predicted outcomes are shown in Figure 1. The results display varying levels of agreement, generally poor to moderate. It is noticeable that there is the least agreement between the top features obtained with Shap and DTD explanation method in for RA30. Table 1 lists the names of the top 5 features obtained by each explanation method for each dataset. For RA30, L1 top 5 predictors list includes one of the predictors identified by senior health care administrators, i.e. previous inpatient visit counts, previous emergency visit counts (stays and presentations), age, index of socioeconomic status and indicator for a stay longer than a day [10]. On the contrary, explanations obtained by Shap and DTD include at least three of the expert-suggested predictors each. It was surprising that the modelling approach which is most accepted by the medical community (i.e. regression) included only one of the predictors suggested by experts. One of the reasons for the unexpected result with L1 could be the particularity of the cohort. An agreement analysis for adult cohorts is needed for more solid and general conclusions on patient readmission problems. However, all three methods agreed on the top feature.

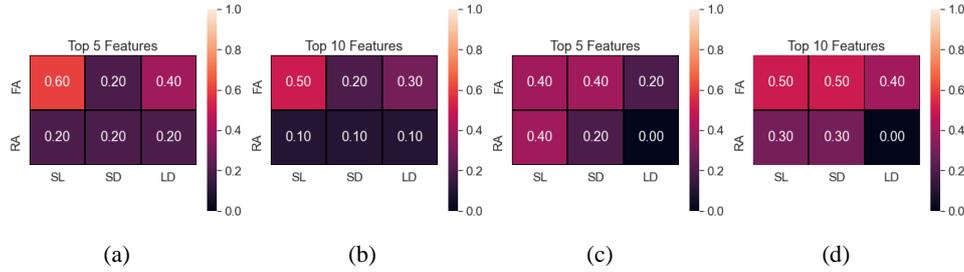

**Figure 1.** Feature Agreement (FA) and Rank Agreement (RA) obtained for Shap - L1 coefficient (SL), Shap-Deep Taylor Decomposition (SD) and L1 coefficient - Deep Taylor Decomposition (LD) model explanations using RA30 (a, b) and VA (c, d) datasets i.e., the outcomes of interest.

**Table 1** Top 5 features obtained with considered explainers for RA30 and VS study cases.

| Study Case | Rank | Shap | L1 coefficients | DTD |
|---|---|---|---|---|
| **RA30** | 1 | Prev. inpat. stay count | Prev. input. stay count | Prev. inpat. stay count |
| | 2 | Patho.: No unique tests | Prev. inpat. stay count$^2$ | Ed presentations stay counts |
| | 3 | Elect. status: Not assigned | Patho.: No unique tests | Age |
| | 4 | LOS | Adm. source: Broader | Prev. inpat. stay count$^2$ |
| | 5 | Planned the same day | Elect. status: Not assigned | Elect. status: Emergency |
| **VS** | 1 | LOS | LOS | AVPU |
| | 2 | SpO2 | SpO2 | O2 Flow rate |
| | 3 | SBP | SBP count | $N_{MesuredEvents}$ |
| | 4 | $N_{RecordedVS}$ | Resp. Rate count | $N_{RecordedVS}$ |
| | 5 | min SpO2 | DBP | SpO2 |

To anticipate patient deterioration from a clinical perspective the influential predictors among the input features align with the observable important clinical vital signs and parameters that experienced clinicians recognize: AVPU (level of consciousness), respiratory function (SpO2, resp Rate and Oxygen therapy), cardiovascular function (SBP systolic and diastolic BP); in addition to the metadata of bedside activity, indicating higher levels of existing care (number of recorded vital signs and bedside events logged, and a baseline measure of chronicity of the acute process (current inpatient length of stay). As the ranking of only the top 5 predictors is considered, the concordance is reasonable but inconsistent across the assessed methods.

## 4. Discussion

From a clinical perspective, 3 sources of discrepancy and or incorrectness were identified. Firstly, incomplete information and missing features lead to failure in explaining all the variations and causation of the outputs. Secondly, dependence between factors and interrelationships with other input features and observations, (e.g. in physiological signals) would lead to different groups of features being picked by different models. Finally, errors and contradictory information could lead to a discrepancy between various outputs.

From a modelling perspective, the discrepancy can be attributed to the optimization objective which is simply the minimisation of error. Consequently, causal and/or clinically relevant associations might be partially or completely missed. Aas et al. [12] have also shown that Shap [7] may lead to incorrect explanations when features are highly correlated. Further, since each of the ML methods operates on different principles, the features identified by the explainability approaches acting post-hoc on these model outputs may vary.

For explainability methods to be trusted and accepted for guiding clinical intervention, sufficient disparate XAI methods would need to agree on influential relationships, changes in observable and input factors would need to cause clinically appropriate changes in model output, and ML model outcomes would need to concur and agree with the real-world results.

## 5. Conclusion

In this work the agreement between explainability methods developed for predictive models in healthcare was investigated. The results indicate that the level of agreement for the main influential features can be poor or misaligned with expert knowledge. Possible causes of disagreement were identified and discussed as the first step towards establishing the criteria for their trustworthiness.